\documentclass[final,5p,times,twocolumn,authoryear]{elsarticle}

\usepackage{amssymb}

\usepackage{amsthm,amsmath}
\RequirePackage{natbib}
\RequirePackage[authoryear]{natbib}                        
\RequirePackage{hyperref}
\usepackage[utf8]{inputenc}                                

\usepackage{amsfonts}                                      
\usepackage{commath}                                       
\usepackage{booktabs}                                      

\usepackage{graphicx}                                      
\usepackage{tikz}                                          
\usepackage{pgf, pgfsys, pgffor}
\usepackage{pgfplots}

\journal{Computational Biology and Chemistry}

\begin{document}

\begin{frontmatter}



\title{Comparison among dimensionality reduction techniques based on Random Projection for cancer classification}


\author[hit]{Haozhe Xie}

\author[hit]{Jie Li\corref{cor}}
\ead{jieli@hit.edu.cn}

\author[hit]{Qiaosheng Zhang}

\author[hit]{Yadong Wang}

\cortext[cor]{Corresponding author}

\address[hit]{School of Computer Science and Technology, Harbin Institute of Technology, No. 92 Xidazhi Street, Harbin 150001, China}

\begin{abstract}
Random Projection (RP) technique has been widely applied in many scenarios because it can reduce high-dimensional features into low-dimensional space within short time and meet the need of real-time analysis of massive data. There is an urgent need of dimensionality reduction with fast increase of big genomics data. However, the performance of RP is usually lower. We attempt to improve classification accuracy of RP through combining other dimension reduction methods such as Principle Component Analysis (PCA), Linear Discriminant Analysis (LDA), and Feature Selection (FS). We compared classification accuracy and running time of different combination methods on three microarray datasets and a simulation dataset. Experimental results show a remarkable improvement of 14.77\% in classification accuracy of FS followed by RP compared to RP on BC-TCGA dataset. LDA followed by RP also helps RP to yield a more discriminative subspace with an increase of 13.65\% on classification accuracy on the same dataset. FS followed by RP outperforms other combination methods in classification accuracy on most of the datasets.
\end{abstract}

\begin{keyword}



Random Projection \sep Dimensionality reduction \sep Classification \sep Breast cancer

\end{keyword}

\end{frontmatter}


\section{Introduction}
\label{introduction}

Machine learning and data mining techniques have been widely applied in many areas \cite{berry1997data, schmidt2015data, han2011data}. The data has very high dimensionality in some of these application scenarios. For example, there are usually up to thousands of merchandises in a hypermarket. Obviously, the market basket data is very high-dimensional data. In text mining, documents are usually represented by a matrix whose dimensionality is equal to the vocabulary size. In these cases, the dimensionality of these matrices is very high due to either a considerable number of merchandises or a wealth of vocabulary. In addition to the high dimensionality, they are often sparse. For gene expression data with several thousands of genes and hundreds of samples, it can be considered as a 2D matrix with continuous value. Unlike the above-mentioned matrices, gene expression matrix is not sparse. High dimensionality may cause ``curse of dimensionality'' which results in inaccurate distance metrics and impact on classification precision. Traditional dimensionality reduction techniques include: Principle Component Analysis (PCA) \cite{jolliffe2002principal}, Linear Discriminant Analysis (LDA) \cite{scholkopft1999fisher} and Feature Selection (FS) based on statistical test \cite{guyon2003introduction, li2007new}. PCA finds dimensions with maximum covariance in an unsupervised way, while LDA tries to seek the best linear discriminant function in a supervised way. FS is based on statistical test that selects features that are instructive in classification. Wrapper method for feature selection \cite{kohavi1997wrappers} trains a new model for classification for each subset, which can usually provide the best performing set for that particular type of model. In spite of this fact, direct use of PCA, LDA and FS may be problematic because these methods require impracticably large computational resources and cannot meet the needs of real-time processing. To address this problem, Random Projection (RP) was purposed in  \cite{dasgupta2000experiments}, which projects original data onto a low-dimensional subspace randomly that the discriminative information can be approximately retained,  and have been developing rapidly these years \cite{zhao2015semi, zhao2016efficient, geppert2015random}. RP has attracted great deal of attention and has been successfully applied to solve large-scale or high-dimensional data analytics, such as classification \cite{zhao2015semi, arriaga2015visual, cannings2015random}, clustering \cite{zhao2016efficient, tasoulis2014random, fern2003random}, regression \cite{geppert2015random, maillard2012linear}, manifold learning \cite{alavi2014random, freund2007learning}, and information retrieval \cite{thaper2002dynamic}. Compared to PCA, LDA and FS, RP is much less expensive in computational cost. However, RP may not capture task-related information because latent space is generated without considering the structure of original data, and accuracy of classifiers based on RP is usually low. So we try to combine RP with other methods to reduce data dimensionality with the purpose of achieving a balance between computational complexity and classification accuracy.

In this paper, we combined RP with PCA, LDA and FS to find a low-dimensional subspace with better discriminative features to classify breast cancer datasets which have a huge amount of genes and small size samples. This paper is organized as follows. In Section \ref{related-work}, the related work is briefly  introduced. The details of methods and experiments are described in Section \ref{combination-methods} and \ref{experiments} respectively. Finally, Conclusion and future work is given in Section \ref{conclusions}.

\section{Related Work}
\label{related-work}

Assuming that the data is represented by a matrix $\mathbf{X} \in \mathbb{R}^{n \times d}$, where $n$ and $d$ are the number of samples and dimension of dataset, and the dimension after reduction is $k$, where $k \ll d$. Dimensionality reduction maps original set of $n$ $d$-dimensional observations to a $k$-dimensional subspace:

\begin{equation}
\widehat{\mathbf{X}}_{n \times k} = \mathbf{X}_{n \times d} \mathbf{W}_{d \times k}
\label{eq:transformation-matrix}
\end{equation}
where $\widehat{\mathbf{X}}_{n \times k}$ indicates the subspaces after reduction, and $\mathbf{W}_{d \times k}$ is the linear transformation matrix. Linear dimensionality reduction approaches are performed by their ways with different transformation matrix $\mathbf{W}$. In the following subsections, four dimensionality reduction methods used in this paper are briefly introduced. 

\subsection{Principle Component Analysis}

PCA finds dimensions with maximum covariance in an unsupervised way \cite{jolliffe2002principal}. It's mathematically defined as an orthogonal linear transformation to convert possibly correlated variables into values of linearly uncorrelated variables. The optimization problem for PCA can be summarized as follows:

\begin{equation}
\mathbf{W}^* = \operatorname*{argmax}\limits_{\mathbf{W} \in \mathbb{R}^{d \times k}} \abs{\frac{\mathbf{W}^T\mathbf{S}\mathbf{W}}{\mathbf{W}^T\mathbf{W}}}
\label{eq:pca-matrix}
\end{equation}
where $\mathbf{S}$ can be defined as follows:

\begin{equation}
\mathbf{S} = \frac{1}{n} \sum_{i=1}^{n} (\mathbf{x}_i - \bar{\mathbf{x}})(\mathbf{x}_i - \bar{\mathbf{x}})^T
\end{equation}
where $\mathbf{x}_i$ is a vector of $d$ dimensions and $\bar{\mathbf{x}}$ represents the mean vector of all samples.

\subsection{Linear Discriminant Analysis}

LDA tries to seek the best linear discriminant function in a supervised way \cite{scholkopft1999fisher}. Compared to PCA, LDA explicitly attempts to model the difference between the classes of data. LDA projects original data onto subspace which has sufficient discriminant power. The dimension of the subspace is restricted to be less than the number of classes. In particular, the dimension of subspace for binary classification problem is 1.

The optimization problem for LDA is described as follows: 

\begin{equation}
\mathbf{W}^* = \operatorname*{argmax}\limits_{\mathbf{W} \in \mathbb{R}^{d \times k}} \abs{\frac{\mathbf{W}^T\mathbf{S}_B\mathbf{W}}{\mathbf{W}^T\mathbf{S}_W\mathbf{W}}}
\label{eq:lda-matrix}
\end{equation}
where $\mathbf{S}_B$ and $\mathbf{S}_W$ are given as follows:

\begin{equation}
\begin{cases}
    \mathbf{S}_B = \sum_{c=1}^{Nc} n_c(\bar{\mathbf{x}}_c-\bar{\mathbf{x}})(\bar{\mathbf{x}}_c-\bar{\mathbf{x}})^T \\
    \mathbf{S}_W = \sum_{i=1}^{n} (\mathbf{x}_i-\bar{\mathbf{x}}_{c_i})(\mathbf{x}_i-\bar{\mathbf{x}}_{c_i})^T
\end{cases}
\end{equation}
where $N_c$ and $n_c$ are the number of classes and the number of samples in $c$-th class respectively, $\bar{\mathbf{x}}$ and $\bar{\mathbf{x}}_c$ denote the mean vector of all samples and the $c$-th class correspondingly, and $c_i$ represents the label of the $i$-th sample.

\subsection{Feature Selection}

There are several statistical tests to select features \cite{li2007new}, such as rank sum test \cite{wilcoxon1945individual}, t-test \cite{student1908probable} and so on. In the paper these methods are called feature selection methods.  Here we take t-test as a representative to study the performance of FS methods derived from statistical test. The t-test assesses whether the means of two groups are statistically different from each other. Normalized distance between two classes of samples can be obtained using the sample mean values: $m_c$, $m_t$ and the sample variances $s_c^2$ and $s_t^2$:

\begin{equation}
t = \frac{m_c - m_t}{\sqrt{\frac{s_c^2}{n_c}-\frac{s_t^2}{n_t}}}
\end{equation}
where $n_c$ and $n_t$ denote the number of two classes of samples respectively, $m=\frac{1}{n} \sum_i x_i$ and $s^2= \frac{1}{n} \sum_i (x_i - m)^2$ are the mean and variance of each sample respectively. Therefore, $t$ follows a t-test with $f$ degree of freedom:

\begin{equation}
f = \frac{[(s_c^2/n_c) + (s_t^2/n_t)]^2}{\frac{(s_c^2/n_c)^2}{n_c-1}+\frac{(s_t^2/n_t)^2}{n_t-1}}
\end{equation}

Two classes of samples are considered significantly different if t exceeds a certain threshold of the selected confidence interval.

The $i$-th sample is denoted by $x_i (i=1, 2, \dots, d)$, and the corresponding label is $y_i \in \left\{0,1\right\}$. To simplify the problem, we assume $x_i (i=1, 2, \dots, d)$ is independent. In this paper, $k$ features with significant differences are selected to build a classifier. The optimal transformation vector $\mathbf{W}$ in Eq. \ref{eq:transformation-matrix} can be formed as follows:

\begin{equation}
\mathbf{W} = \begin{bmatrix}
    w_{1, 1} & w_{1, 2} & \dots  & w_{1, k} \\
    w_{2, 1} & w_{2, 2} & \dots  & w_{2, k} \\
    \vdots   & \vdots   & \ddots & \vdots \\
    w_{d, 1} & w_{d, 2} & \dots  & w_{d, k} \\
\end{bmatrix}
\label{eq:fs-matrix}
\end{equation}
where
\begin{equation}
\mathbf{w}_{i, j} = 
\begin{cases}
1 & \text{i-th feature is selected} \\
0 & \text{i-th feature is not selected}
\end{cases}
\end{equation}

\subsection{Random Projection}

In  \cite{bingham2001random} and  \cite{li2006very}, RP has been proposed to address the burdensome computation time in dimensionality reduction. In RP, transformation matrix $\mathbf{W}$ is randomly generated. Experiment results  \cite{bingham2001random, li2006very} have reveals that RP is computationally less expensive with a little distortion of original data. On the other hand, RP may not capture intrinsic structure underlying original data because $\mathbf{W}$ is generated without considering the structure of data.

As suggested in  \cite{li2006very}, the linear transform matrix $\mathbf{W}$ is sampled as follows:

\begin{equation}
w_{i, j} = \sqrt{c} \begin{cases}
1  & \text{with \ prob.} \frac{1}{2c} \\
0  & \text{with \ prob.} 1 - \frac{1}{c} \\
-1 & \text{with \ prob.} \frac{1}{2c} \\
\end{cases}
\label{eq:rp-matrix}
\end{equation}
where $c$ is set to $\sqrt{d}$.

\section{Combination Methods Based on Random Projection}
\label{combination-methods}

In this section, RP is combined with other methods, such as PCA, LDA and FS to improve the performance of RP.

\subsection{Random Projection + Principle Component Analysis (RP+PCA)}

PCA seeks direction with maximum variance. We wonder whether RP based on PCA can improve classification accuracy.

Assume that matrix $\mathbf{X} \in \mathbb{R}^{n \times d}$ is consisted of n samples and d dimension. Firstly, RP is used to map original data to a subspace $\widehat{\mathbf{X}_1} \in \mathbb{R}^{n \times k_1}$ by a random matrix $\mathbf{W}_1 \in \mathbb{R}^{d \times k_1}$ according to Eq. \ref{eq:rp-matrix}. Then PCA projects $\widehat{\mathbf{X}_1} \in \mathbb{R}^{n \times k_1}$ to $\widehat{\mathbf{X}_2} \in \mathbb{R}^{n \times k_2}$ by a matrix $\mathbf{W}_2 \in \mathbb{R}^{k_1 \times k_2}$ from Eq. \ref{eq:pca-matrix}.

\subsection{Random Projection + Linear Discriminant Analysis (RP+LDA)}

Here we want to know whether LDA is benefit for RP. RP is combined with LDA which is a supervised approach to find the best linear discriminant function.

Suppose data is described by a matrix $\mathbf{X} \in \mathbb{R}^{n \times d}$. First RP maps original data to a subspace $\widehat{\mathbf{X}_1} \in \mathbb{R}^{n \times k_1}$ by a random matrix $\mathbf{W}_1 \in \mathbb{R}^{d \times k_1}$ announced in Eq. \ref{eq:rp-matrix}. Then LDA coverts $\widehat{\mathbf{X}_1} \in \mathbb{R}^{n \times k_1}$ to $\widehat{\mathbf{X}_2} \in \mathbb{R}^{n \times k_2}$ by a linear transformation matrix $\mathbf{W}_2 \in \mathbb{R}^{k_1 \times k_2}$ according to Eq. \ref{eq:lda-matrix}.

\subsection{Random Projection + Feature Selection (RP+FS)}

In subspaces generated by RP, some dimensions don't have significant differences, and this may affect the  accuracy in classification. We believe FS can help to filter them.

According to Eq. \ref{eq:rp-matrix}, a subset $\widehat{\mathbf{X}_1} \in \mathbb{R}^{n \times k_1}$ with $k_1$ features is generated in a random way. Later, FS projects data onto $\widehat{\mathbf{X}_2} \in \mathbb{R}^{n \times k_2}$ by a matrix $\mathbf{W}_2 \in \mathbb{R}^{k_1 \times k_2}$ declared in Eq. \ref{eq:fs-matrix}.

\subsection{Feature Selection + Random Projection (FS+RP)}

In this method, FS is used before RP to filter dimensions that are similar between two classes of samples in original dataset. The preprocessing conducted by FS may help RP yield more discriminant subspaces.

Matrix $\mathbf{W}_1 \in \mathbb{R}^{d \times k_1}$ with $k_1$ features is firstly calculated by FS with a matrix described in Eq. \ref{eq:fs-matrix}. Another matrix $\mathbf{W}_2 \in \mathbb{R}^{k_1 \times k_2}$ is calculated using Eq. \ref{eq:rp-matrix} in order to obtain a subspace $\widehat{\mathbf{X}_2} \in \mathbb{R}^{n \times k_2}$ by RP.

\subsection{Feature Selection + Random Projection + Linear Discriminant Analysis (FS+RP+LDA)}

Here three methods: FS, RP and LDA are combined to reduce dimensionality. Compared with FS+RP, LDA is added to study whether LDA is useful to improve the subspace generated by FS+RP.

This method includes three steps: First of all, FS transforms original data into $\widehat{\mathbf{X}_1} \in \mathbb{R}^{n \times k_1}$ with a matrix $\mathbf{W}_1 \in \mathbb{R}^{d \times k_1}$ stated in Eq. \ref{eq:fs-matrix}. Secondly, RP is employed to generate a linear transformation matrix $\mathbf{W}_2 \in \mathbb{R}^{k_1 \times k_2}$  whose values are random assigned and maps data to a subspace $\widehat{\mathbf{X}_2} \in \mathbb{R}^{n \times k_2}$ with $k_2$ dimension. Finally, LDA is taken to form $\widehat{\mathbf{X}_3} \in \mathbb{R}^{n \times k_3}$ with a matrix $\mathbf{W}_3 \in \mathbb{R}^{k_2 \times k_3}$ determined in Eq. \ref{eq:lda-matrix}.

\subsection{Feature Selection + Random Projection + Post Feature Selection (FS+RP+PFS)}

In the method, we plan to employ a new method called ``Post Feature Selection'' (PFS) to select features after FS+RP in order to find a set of features which have sufficient discriminative power.

PFS picks column vectors out of $\mathbf{X}$ that have the maximum classification accuracy. $y_i$ denotes the label of the $i$-th sample $x_i$ of $\mathbf{X}$. Training and testing datasets include $m$ and $n$ samples  respectively are randomly selected from $\mathbf{X}$. A Support Vector Machine (SVM) with radial basis kernel function is used to evaluate the classification accuracy for each dimension, where top $k$ dimensions with higher accuracy are selected.

This method firstly derives a subspace $\widehat{\mathbf{X}_1} \in \mathbb{R}^{n \times k_1}$ using a matrix $\mathbf{W}_1 \in \mathbb{R}^{d \times k_1}$ announced in Eq. \ref{eq:fs-matrix}. A subset $\widehat{\mathbf{X}_2} \in \mathbb{R}^{n \times k_2}$ with $k_2$ features is calculated in a random way with $\mathbf{W}_2 \in \mathbb{R}^{k_1 \times k_2}$ in the next stage. Finally, PFS is adopted to generate $\mathbf{W}_3 \in \mathbb{R}^{k_2 \times k_3}$ and corresponding subspace $\widehat{\mathbf{X}_3} \in \mathbb{R}^{n \times k_3}$, where $k_3$ is the number of features selected in PFS.

\section{Experiments}
\label{experiments}

\subsection{Datasets and Experimental Environment}

In this section, the performance of methods described above was tested and compared on three breast cancer gene expression data \cite{cancer2012comprehensive, wang2005gene, hatzis2011genomic}, which can be downloaded from https://tcga-data.nci.nih.gov/tcga and https://www.ncbi.nlm.nih.gov/geo. Breast cancer from TCGA (BC-TCGA) \cite{cancer2012comprehensive} consists of 17814 genes and 590 samples (including 61 normal tissue samples and 529 breast cancer tissue samples). GSE2034 \cite{wang2005gene} includes 12634 genes and 286 breast cancer samples (including 107 recurrence tumor samples and 179 no recurrence samples). GSE25066 \cite{hatzis2011genomic} has 492 breast cancer samples available (including 100 pathologic complete response (PCR) samples and 392 residual disease (RD) samples) and 12634 genes. Some missing values in these datasets were filled with gene mean values of each group and expression values of each gene were processed by Z-score. The above methods were implemented with scikit-learn\footnote{http://scikit-learn.org/stable/}, which is one of the most popular packages for machine learning in Python. The environments for all methods performed in our experiments are as follows: a Linux PC running with Intel Xeon E5504 (2.00 GHz) processor and 24 GB RAM. In order to test the classification accuracy and running time of different methods, a SVM was employed to classify two classes of samples. Thus, methods' running time mainly includes two parts: one is the time spending on reducing dimensionality; the other is the time used for training a SVM.

\subsection{Experimental Procedure}

Here we take RP+PCA as an example to introduce the detailed procedure of the whole experiment on BC-TCGA. First, we get training and testing sets. Training sample set includes 30 normal and 30 tumor tissue samples randomly selected from BC-TCGA without replacement, and testing sample set includes 30 normal and 30 tumor tissue samples randomly selected from remaining samples. Then, the training dataset is mapped into a subspace $\widehat{\mathbf{X}'}_{\rm Train}$ by a random matrix $\mathbf{W}_{\rm RP}$. Then, a linear matrix $\mathbf{W}_{\rm PCA}$ derived from PCA projects $\widehat{\mathbf{X}'}_{\rm Train}$ onto another subspace $\widehat{\mathbf{X}}_{\rm Train}$. Next, $\mathbf{W}_{\rm RP}$ and $\mathbf{W}_{\rm PCA}$ are applied to test dataset to obtain $\widehat{\mathbf{X}}_{\rm Test}$. Finally, a SVM with radial basis kernel function is trained and tested on $\widehat{\mathbf{X}}_{\rm Train}$ and $\widehat{\mathbf{X}}_{\rm Test}$ respectively. When the number of spatial dimension changes, classification accuracy and running time of SVMs also change for all methods described above. In order to investigate the effect of change in dimension, we conducted a sequence of experiments in different dimensions: 10, 50, 100, 200, 400, 800, 1000, 1250, 1500, 1750, 2000, 2500, 3000, 4000, and 5000. The above procedure is repeated 100 times. Other methods also follow a similar process on these datasets, except for the difference that training and testing sets are consisted of 50 positive samples and 50 negative samples. Positive samples stand for no recurrence samples in GSE2034 and PCR samples in GSE25066, while negative samples refer to recurrence samples in GSE2034 and RD samples in GSE25066. Finally, average values of classification accuracy and corresponding running time are obtained.

\begin{table*}[!htb]
    \caption{Highest classification accuracy (\%) on three gene expression datasets}
    \begin{tabular*}{\textwidth}{l @{\extracolsep{\fill}} rrrrrr}
    \toprule
                       & \multicolumn{2}{c}{\textbf{BC-TCGA}} & \multicolumn{2}{c}{\textbf{GSE2034}} & \multicolumn{2}{c}{\textbf{GSE25066}} \\
    \midrule
    \textbf{Methods}   & Training Set      & Testing Set      & Training Set      & Testing Set      & Training Set      & Testing Set \\
    \midrule
    RP                 & 100.00            & 86.23            & 97.85             & 59.59            & 100.00            & 66.90 \\
    FS + RP            & 99.95             & \textbf{98.97}   & 99.88             & 61.25            & 96.48             & \textbf{71.07} \\
    FS + RP + LDA      & 99.93             & 98.67            & 80.71             & 60.56            & 87.81             & 69.56 \\
    FS + RP + PFS      & 99.97             & 98.95            & 99.99             & \textbf{61.55}   & 68.50             & 67.06 \\
    RP + FS            & 100.00            & 92.70            & 100.00            & 60.89            & 99.98             & 68.65 \\ 
    RP + PCA           & 100.00            & 76.70            & 100.00            & 54.25            & 100.00            & 60.44 \\
    RP + LDA           & 98.68             & 98.00            & 73.65             & 58.87            & 80.23             & 69.21 \\
    \bottomrule
    \multicolumn{7}{l}{* Bold face indicates the highest classification accuracy on test data}
    \end{tabular*}
\label{tab:accuracy-comparison-on-gene-data}
\end{table*}

\subsection{Experimental Results}

\begin{figure*}[!htb]
\centering
\resizebox{.6\linewidth}{!} {
    \begin{tikzpicture}
    \begin{axis}[
        xbar stacked,
        xlabel            = {Running Time (s)},
        legend style      = {at = {(0.5,-0.15)}, anchor = north, legend columns = -1, column sep = 0.25cm},
        legend cell align = left,
        xmajorgrids       = true,
        stack negative    = separate,
        xtick             = {-7, -6, -5, -4, -3, -2, -1, 0, 1, 2, 3, 4, 5, 6, 7},
        xticklabels       = {, , , Offline, , , , 0, 1, 2, 3, 4, 5, 6, 7},
        ytick             = {0, 1, 2, 3, 4, 5, 6, 7, 8, 9, 10, 11, 12, 13, 14, 15, 16, 17, 18, 19, 20, 21, 22, 23, 24, 25},
        yticklabels       = {\textbf{Dimension = 150}, RP, FS+RP, FS+RP+LDA, FS+RP+PFS, RP+FS, RP+PCA, RP+LDA, , \textbf{Dimension = 1000}, RP, FS+RP, FS+RP+LDA, FS+RP+PFS, RP+FS, RP+PCA, RP+LDA, , \textbf{Dimension = 3000}, RP, FS+RP, FS+RP+LDA, FS+RP+PFS, RP+FS, RP+PCA, RP+LDA},
        bar width         = 2mm,
        width             = \textwidth,
        y dir             = reverse,
    ]
    
    \addplot[
        fill=lightgray,
    ]
    coordinates {
        (0, 0)(0, 1)(-6.770412, 2)(-6.815493, 3)(-6.921003, 4)(0, 5)(0, 6)(0, 7)(0, 8)(0, 9)(0, 10)(-6.795959, 11)(-6.82081, 12)(-6.866169, 13)(0, 14)(0, 15)(0, 16)(0, 17)(0, 18)(0, 19)(-6.894662, 20)(-6.896096, 21)(-6.847853, 22)(0, 23)(0, 24)(0, 25)
    };
    \addlegendentry{Offline FS};
    
    \addplot[
        fill=orange,
    ]
    coordinates {
        (0, 0)(0.716926, 1)(0.021, 2)(0.0205164, 3)(0.020431, 4)(1.390247, 5)(1.556214, 6)(0.726104, 7)(0, 8)(0, 9)(0.833121, 10)(0.021, 11)(0.145571, 12)(0.149831, 13)(1.388866, 14)(1.552901, 15)(1.007191, 16)(0, 17)(0, 18)(1.219722, 19)(0.683791, 20)(0.683882, 21)(0.685092, 22)(1.399184, 23)(1.554499, 24)(1.391188, 25)
    };
    \addlegendentry{RP};
    
    \addplot[
        fill=cyan,
    ]
    coordinates {
        (0, 0)(0, 1)(0, 2)(0, 3)(0, 4)(2.908382, 5)(0, 6)(0, 7)(0, 8)(0, 9)(0, 10)(0, 11)(0, 12)(0, 13)(3.13971, 14)(0, 15)(0, 16)(0, 17)(0, 18)(0, 19)(0, 20)(0, 21)(0, 22)(3.13971, 23)(0, 24)(0, 25)
    };
    \addlegendentry{FS};
    
    \addplot[
        fill=red,
    ]
    coordinates {
        (0, 0)(0, 1)(0, 2)(0, 3)(0, 4)(0, 5)(0.144723, 6)(0, 7)(0, 8)(0, 9)(0, 10)(0, 11)(0, 12)(0, 13)(0, 14)(0.144495, 15)(0, 16)(0, 17)(0, 18)(0, 19)(0, 20)(0, 21)(0, 22)(0, 23)(0.14376, 24)(0, 25)
    };
    \addlegendentry{PCA};
    
    \addplot[
        fill=darkgray,
    ]
    coordinates {
        (0, 0)(0, 1)(0, 2)(0.014646, 3)(0, 4)(0, 5)(0, 6)(0.014765, 7)(0, 8)(0, 9)(0, 10)(0, 11)(0.047575, 12)(0, 13)(0, 14)(0, 15)(0.122673, 16)(0, 17)(0, 18)(0, 19)(0, 20)(0.127439, 21)(0, 22)(0, 23)(0, 24)(0.122673, 25)
    };
    \addlegendentry{LDA};
    
    \addplot[
        fill=teal,
    ]
    coordinates {
        (0, 0)(0, 1)(0, 2)(0, 3)(0.280686, 4)(0, 5)(0, 6)(0, 7)(0, 8)(0, 9)(0, 10)(0, 11)(0, 12)(1.903888, 13)(0, 14)(0, 15)(0, 16)(0, 17)(0, 18)(0, 19)(0, 20)(0, 21)(5.773145, 22)(0, 23)(0, 24)(0, 25)
    };
    \addlegendentry{PFS};
    
    \addplot[
        fill=purple,
    ]
    coordinates {
        (0, 0)(0.011751, 1)(0.011204, 2)(0.002303, 3)(0.009232, 4)(0.019995, 5)(0.009232, 6)(0.002121, 7)(0, 8)(0, 9)(0.049416, 10)(0.04899, 11)(0.002387, 12)(0.084795, 13)(0.104008, 14)(0.009349, 15)(0.002751, 16)(0, 17)(0, 18)(0.143874, 19)(0.143898, 20)(0.002806, 21)(0.196272, 22)(0.28884, 23)(0.009323, 24)(0.002677, 25)
    };
    \addlegendentry{Classification};
    \end{axis}  
    \end{tikzpicture}
}
\caption{Running time of different methods on GSE25066}
\label{fig:running-time-on-gse25066}
\end{figure*}
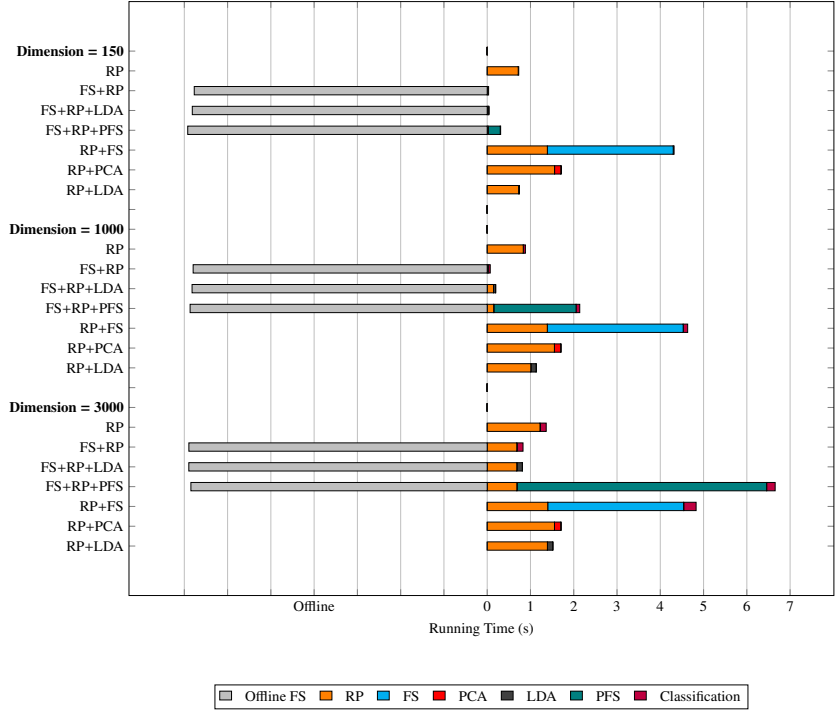

The experiment results on these gene expression datasets are shown in Table \ref{tab:accuracy-comparison-on-gene-data} and Figure \ref{fig:running-time-on-gse25066}. Classification accuracy of different methods on the three datasets is shown in Table \ref{tab:accuracy-comparison-on-gene-data}. It can be seen that accuracy of dimensionality reduction methods on these datasets is different. On the first dataset all methods get high classification accuracy on testing set, while on the other two datasets obtain relatively low accuracy. This indicates that two classes of samples in GSE2034 and GSE25066 seem to be more difficult to classify.

\begin{figure*}[!htb]
\centering
\resizebox{.6\linewidth}{!} {
    \begin{tikzpicture}[every plot/.append style={ultra thick}]
    \begin{axis} [
        width             = \textwidth,
        xlabel            = {Number of Spatial Dimensions or Features},
        ylabel            = {Classification Accuracy (\%)},
        xtick             = {0, 1, 2, 3, 4, 5, 6, 7, 8, 9, 10, 11, 12, 13, 14},
        xticklabels       = {10, 50, 100, 200, 400, 800, 1000, 1250, 1500, 1750, 2000, 2500, 3000, 4000, 5000},
        xticklabel style  = {font=\fontsize{10}{10}\selectfont},
        ytick             = {50, 55, 60, 65, 70, 75},
        legend style      = {at = {(0.5,-0.1)}, anchor = north, legend columns = 4, column sep = 0.25cm},
        legend cell align = left,
        ymajorgrids       = true,
        grid style        = dashed,
    ]

    \addplot[
        color=purple,
        mark=square,
    ]
    coordinates {
        (0, 52.34)(1, 56.20)(2, 50.48)(3, 50.67)(4, 52.47)(5, 54.11)(6, 56.37)(7, 56.55)(8, 57.18)(9, 59.30)(10, 60.86)(11, 63.18)(12, 65.81)(13, 66.31)(14, 66.90)
    };
    \addlegendentry{RP};
    
    \addplot[
        color=orange,
        mark=triangle,
    ]
    coordinates {
        (0, 65.34)(1, 71.07)(2, 69.88)(3, 70.52)(4, 70.87)(5, 70.55)(6, 70.16)(7, 70.11)(8, 69.98)(9, 69.80)(10, 70.36)(11, 70.01)(12, 70.18)(13, 69.97)
    };
    \addlegendentry{FS+RP};
    
    \addplot[
        color=teal,
        mark=*,
    ]
    coordinates {
        (0, 68.35)(1, 60.33)(2, 50.62)(3, 62.72)(4, 67.24)(5, 68.90)(6, 69.37)(7, 69.56)(8, 69.42)(9, 68.95)(10, 69.43)(11, 69.38)(12, 68.88)(13, 68.57)
    };
    \addlegendentry{FS+RP+LDA};
    
    \addplot[
        color=darkgray,
        mark=square*,
    ]
    coordinates {
        (0, 64.70)(1, 65.80)(2, 66.30)(3, 66.21)(4, 65.85)(5, 66.98)(6, 67.06)(7, 66.28)(8, 66.63)(9, 66.38)(10, 66.01)(11, 65.85)(12, 66.00)
    };
    \addlegendentry{FS+RP+PFS};
    
    \addplot[
        color=red,
        mark=triangle*,
    ]
    coordinates {
        (0, 64.73)(1, 68.03)(2, 67.57)(3, 68.65)(4, 67.82)(5, 67.73)(6, 67.74)(7, 68.16)(8, 68.47)(9, 67.76)(10, 68.44)(11, 68.17)(12, 67.91)(13, 67.15)
    };
    \addlegendentry{RP+FS};
    
    \addplot[
        color=violet,
        mark=otimes,
    ]
    coordinates {
        (0, 54.59)(1, 57.29)(2, 59.05)(3, 58.79)(4, 59.80)(5, 59.48)(6, 59.01)(7, 59.32)(8, 58.92)(9, 58.96)(10, 59.25)(11, 60.44)(12, 59.69)(13, 59.88)
    };
    \addlegendentry{RP+PCA};
  
    \addplot[
        color=cyan,
        mark=diamond,
    ]
    coordinates {
        (0, 58.37)(1, 57.09)(2, 50.43)(3, 60.31)(4, 65.08)(5, 68.10)(6, 68.04)(7, 68.88)(8, 68.76)(9, 68.74)(10, 68.21)(11, 69.21)(12, 68.23)(13, 68.52)(14, 68.99)
    };
    \addlegendentry{RP+LDA};
    \end{axis}
    \end{tikzpicture}
}
\caption{Classification accuracy of different methods on the testing set of GSE25066}
\label{fig:classification-accuracy-gse25066}
\end{figure*}
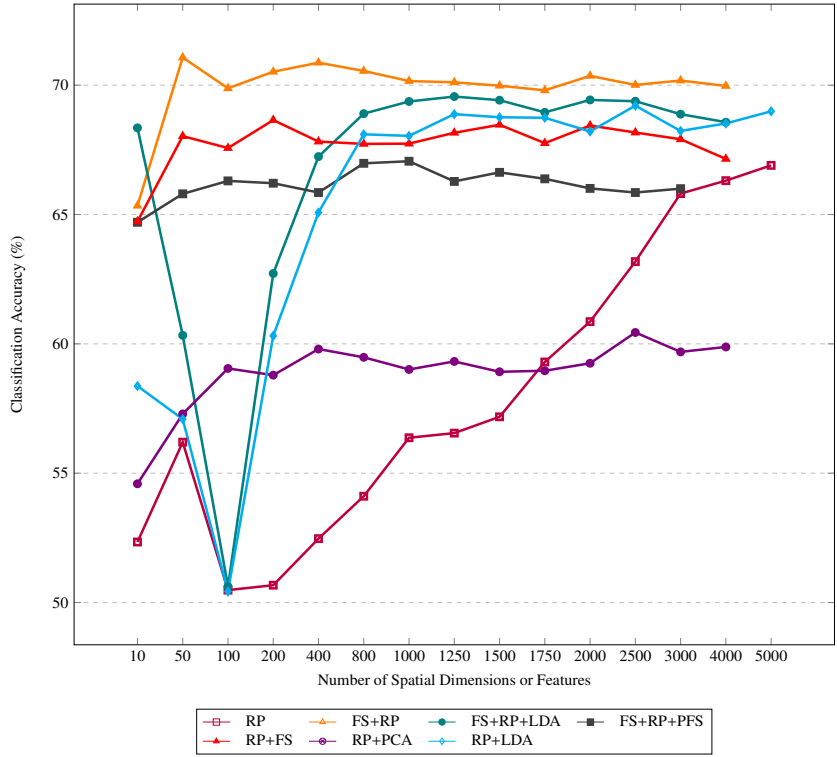

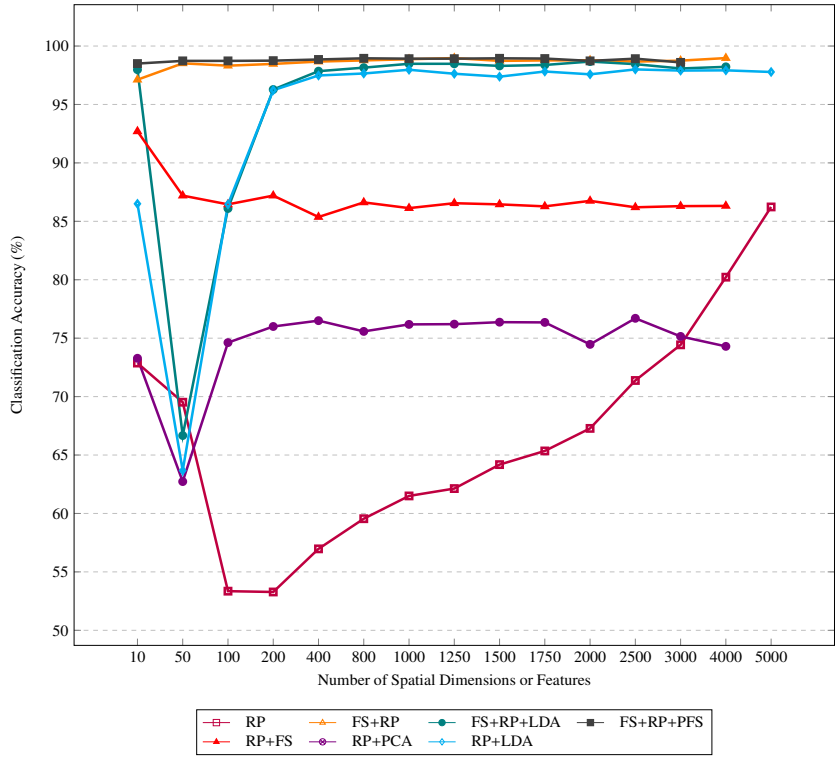
\begin{figure*}[!htb]
\centering
\resizebox{.6\linewidth}{!} {
    \begin{tikzpicture}[every plot/.append style={ultra thick}]
    \begin{axis} [
        width             = \textwidth,
        xlabel            = {Number of Spatial Dimensions or Features},
        ylabel            = {Classification Accuracy (\%)},
        xtick             = {0, 1, 2, 3, 4, 5, 6, 7, 8, 9, 10, 11, 12, 13, 14},
        xticklabels       = {10, 50, 100, 200, 400, 800, 1000, 1250, 1500, 1750, 2000, 2500, 3000, 4000, 5000},
        xticklabel style  = {font=\fontsize{10}{10}\selectfont},
        ytick             = {50, 55, 60, 65, 70, 75, 80, 85, 90, 95, 100},
        legend style      = {at = {(0.5,-0.1)}, anchor = north, legend columns = 4, column sep = 0.25cm},
        legend cell align = left,
        ymajorgrids       = true,
        grid style        = dashed,
    ]

    \addplot[
        color=purple,
        mark=square,
    ]
    coordinates {
        (0, 72.87)(1, 69.52)(2, 53.35)(3, 53.28)(4, 56.97)(5, 59.55)(6, 61.50)(7, 62.13)(8, 64.18)(9, 65.35)(10, 67.28)(11, 71.38)(12, 74.43)(13, 80.22)(14, 86.23)
    };
    \addlegendentry{RP};
    
    \addplot[
        color=orange,
        mark=triangle,
    ]
    coordinates {
        (0, 97.12)(1, 98.52)(2, 98.32)(3, 98.47)(4, 98.67)(5, 98.77)(6, 98.87)(7, 98.97)(8, 98.73)(9, 98.77)(10, 98.78)(11, 98.67)(12, 98.75)(13, 98.97)
    };
    \addlegendentry{FS+RP};
    
    \addplot[
        color=teal,
        mark=*,
    ]
    coordinates {
        (0, 97.98)(1, 66.67)(2, 86.10)(3, 96.28)(4, 97.85)(5, 98.15)(6, 98.48)(7, 98.48)(8, 98.30)(9, 98.38)(10, 98.67)(11, 98.45)(12, 98.08)(13, 98.22)
    };
    \addlegendentry{FS+RP+LDA};
    
    \addplot[
        color=darkgray,
        mark=square*,
    ]
    coordinates {
        (0, 98.50)(1, 98.73)(2, 98.73)(3, 98.75)(4, 98.85)(5, 98.95)(6, 98.92)(7, 98.92)(8, 98.95)(9, 98.93)(10, 98.73)(11, 98.92)(12, 98.60)
    };
    \addlegendentry{FS+RP+PFS};
    
    \addplot[
        color=red,
        mark=triangle*,
    ]
    coordinates {
        (0, 92.70)(1, 87.20)(2, 86.45)(3, 87.20)(4, 85.37)(5, 86.62)(6, 86.13)(7, 86.55)(8, 86.45)(9, 86.28)(10, 86.75)(11, 86.20)(12, 86.30)(13, 86.32)
    };
    \addlegendentry{RP+FS};
    
    \addplot[
        color=violet,
        mark=otimes,
    ]
    coordinates {
        (0, 73.28)(1, 62.73)(2, 74.62)(3, 76.00)(4, 76.50)(5, 75.58)(6, 76.18)(7, 76.20)(8, 76.37)(9, 76.35)(10, 74.47)(11, 76.70)(12, 75.15)(13, 74.30)
    };
    \addlegendentry{RP+PCA};
  
    \addplot[
        color=cyan,
        mark=diamond,
    ]
    coordinates {
        (0, 86.50)(1, 63.67)(2, 86.45)(3, 96.22)(4, 97.48)(5, 97.65)(6, 97.97)(7, 97.63)(8, 97.38)(9, 97.82)(10, 97.58)(11, 98.00)(12, 97.90)(13, 97.92)(14, 97.78)
    };
    \addlegendentry{RP+LDA};
    \end{axis}
    \end{tikzpicture}
}
\caption{Classification accuracy of different methods on the testing set of BC-TCGA}
\label{fig:classification-accuracy-bc-tcga}
\end{figure*}

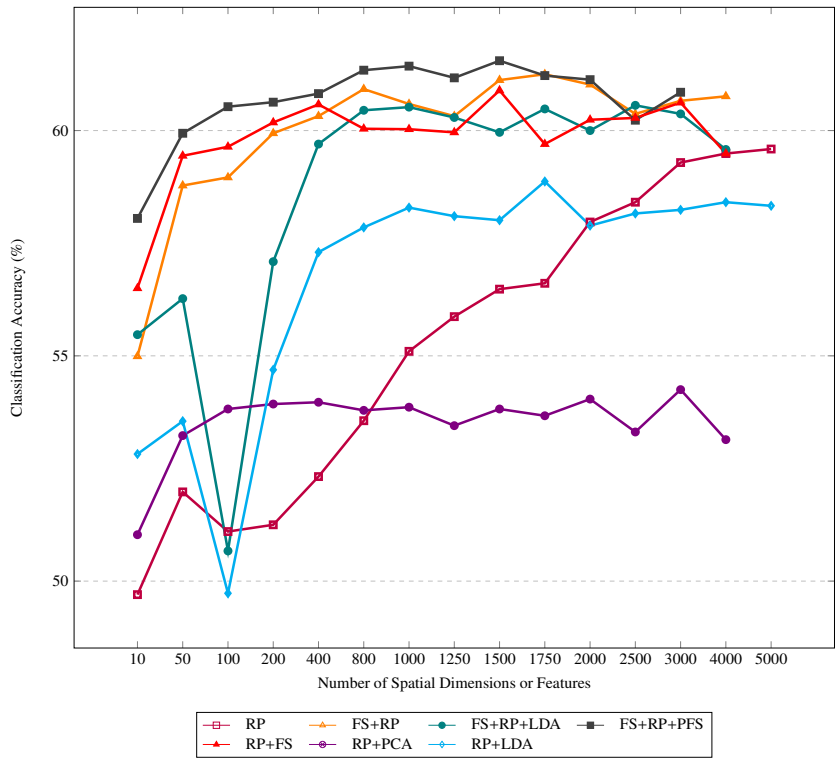
\begin{figure*}[!htb]
\centering
\resizebox{.6\linewidth}{!} {
    \begin{tikzpicture}[every plot/.append style={ultra thick}]
    \begin{axis} [
        width             = \textwidth,
        xlabel            = {Number of Spatial Dimensions or Features},
        ylabel            = {Classification Accuracy (\%)},
        xtick             = {0, 1, 2, 3, 4, 5, 6, 7, 8, 9, 10, 11, 12, 13, 14},
        xticklabels       = {10, 50, 100, 200, 400, 800, 1000, 1250, 1500, 1750, 2000, 2500, 3000, 4000, 5000},
        xticklabel style  = {font=\fontsize{10}{10}\selectfont},
        ytick             = {45, 50, 55, 60, 65},
        legend style      = {at = {(0.5,-0.1)}, anchor = north, legend columns = 4, column sep = 0.25cm},
        legend cell align = left,
        ymajorgrids       = true,
        grid style        = dashed,
    ]

    \addplot[
        color=purple,
        mark=square,
    ]
    coordinates {
        (0, 49.70)(1, 51.98)(2, 51.10)(3, 51.25)(4, 52.32)(5, 53.56)(6, 55.10)(7, 55.87)(8, 56.48)(9, 56.61)(10, 57.97)(11, 58.41)(12, 59.29)(13, 59.49)(14, 59.59)
    };
    \addlegendentry{RP};
    
    \addplot[
        color=orange,
        mark=triangle,
    ]
    coordinates {
        (0, 54.99)(1, 58.78)(2, 58.96)(3, 59.94)(4, 60.32)(5, 60.92)(6, 60.59)(7, 60.32)(8, 61.12)(9, 61.25)(10, 61.02)(11, 60.37)(12, 60.66)(13, 60.76)
    };
    \addlegendentry{FS+RP};
    
    \addplot[
        color=teal,
        mark=*,
    ]
    coordinates {
        (0, 55.47)(1, 56.27)(2, 50.67)(3, 57.09)(4, 59.70)(5, 60.45)(6, 60.52)(7, 60.29)(8, 59.96)(9, 60.48)(10, 60.00)(11, 60.56)(12, 60.37)(13, 59.58)
    };
    \addlegendentry{FS+RP+LDA};
    
    \addplot[
        color=darkgray,
        mark=square*,
    ]
    coordinates {
        (0, 58.05)(1, 59.94)(2, 60.53)(3, 60.63)(4, 60.82)(5, 61.34)(6, 61.43)(7, 61.17)(8, 61.55)(9, 61.22)(10, 61.13)(11, 60.23)(12, 60.85)
    };
    \addlegendentry{FS+RP+PFS};
    
    \addplot[
        color=red,
        mark=triangle*,
    ]
    coordinates {
        (0, 56.50)(1, 59.44)(2, 59.64)(3, 60.18)(4, 60.58)(5, 60.04)(6, 60.03)(7, 59.96)(8, 60.89)(9, 59.70)(10, 60.24)(11, 60.28)(12, 60.62)(13, 59.49)
    };
    \addlegendentry{RP+FS};
    
    \addplot[
        color=violet,
        mark=otimes,
    ]
    coordinates {
        (0, 51.03)(1, 53.23)(2, 53.82)(3, 53.93)(4, 53.97)(5, 53.79)(6, 53.86)(7, 53.45)(8, 53.82)(9, 53.67)(10, 54.04)(11, 53.31)(12, 54.25)(13, 53.14)
    };
    \addlegendentry{RP+PCA};
  
    \addplot[
        color=cyan,
        mark=diamond,
    ]
    coordinates {
        (0, 52.82)(1, 53.55)(2, 49.73)(3, 54.69)(4, 57.30)(5, 57.85)(6, 58.29)(7, 58.10)(8, 58.01)(9, 58.87)(10, 57.89)(11, 58.16)(12, 58.24)(13, 58.41)(14, 58.33)
    };
    \addlegendentry{RP+LDA};
    \end{axis}
    \end{tikzpicture}
}
\caption{Classification accuracy of different methods on the testing set of GSE2034}
\label{fig:classification-accuracy-gse2034}
\end{figure*}

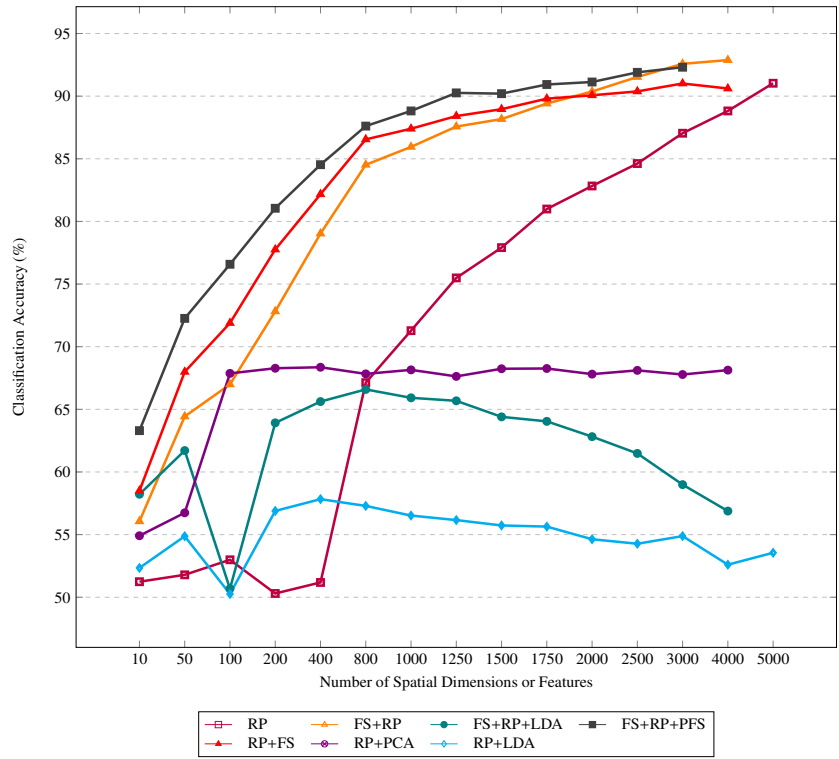
\begin{figure*}
\centering
\resizebox{.6\linewidth}{!} {
    \begin{tikzpicture}[every plot/.append style={ultra thick}]
    \begin{axis} [
        width             = \textwidth,
        xlabel            = {Number of Spatial Dimensions or Features},
        ylabel            = {Classification Accuracy (\%)},
        xtick             = {0, 1, 2, 3, 4, 5, 6, 7, 8, 9, 10, 11, 12, 13, 14},
        xticklabels       = {10, 50, 100, 200, 400, 800, 1000, 1250, 1500, 1750, 2000, 2500, 3000, 4000, 5000},
        xticklabel style  = {font=\fontsize{10}{10}\selectfont},
        ytick             = {45, 50, 55, 60, 65, 70, 75, 80, 85, 90, 95},
        legend style      = {at = {(0.5,-0.1)}, anchor = north, legend columns = 4, column sep = 0.25cm},
        legend cell align = left,
        ymajorgrids       = true,
        grid style        = dashed,
    ]

    \addplot[
        color=purple,
        mark=square,
    ]
    coordinates {
        (0, 51.24)(1, 51.79)(2, 52.99)(3, 50.30)(4, 51.18)(5, 67.13)(6, 71.28)(7, 75.49)(8, 77.91)(9, 80.99)(10, 82.83)(11, 84.62)(12, 87.04)(13, 88.82)(14, 91.03)
    };
    \addlegendentry{RP};
    
    \addplot[
        color=orange,
        mark=triangle,
    ]
    coordinates {
        (0, 56.06)(1, 64.43)(2, 66.99)(3, 72.82)(4, 79.03)(5, 84.52)(6, 85.95)(7, 87.57)(8, 88.17)(9, 89.41)(10, 90.37)(11, 91.53)(12, 92.58)(13, 92.88)
    };
    \addlegendentry{FS+RP};
    
    \addplot[
        color=teal,
        mark=*,
    ]
    coordinates {
        (0, 58.22)(1, 61.71)(2, 50.68)(3, 63.92)(4, 65.62)(5, 66.59)(6, 65.92)(7, 65.68)(8, 64.40)(9, 64.04)(10, 62.82)(11, 61.48)(12, 58.99)(13, 56.88)
    };
    \addlegendentry{FS+RP+LDA};
    
    \addplot[
        color=darkgray,
        mark=square*,
    ]
    coordinates {
        (0, 63.30)(1, 72.26)(2, 76.58)(3, 81.05)(4, 84.53)(5, 87.61)(6, 88.82)(7, 90.26)(8, 90.20)(9, 90.93)(10, 91.13)(11, 91.90)(12, 92.31)
    };
    \addlegendentry{FS+RP+PFS};
    
    \addplot[
        color=red,
        mark=triangle*,
    ]
    coordinates {
        (0, 58.50)(1, 67.99)(2, 71.89)(3, 77.76)(4, 82.17)(5, 86.55)(6, 87.40)(7, 88.41)(8, 88.96)(9, 89.80)(10, 90.07)(11, 90.38)(12, 91.01)(13, 90.61)
    };
    \addlegendentry{RP+FS};
    
    \addplot[
        color=violet,
        mark=otimes,
    ]
    coordinates {
        (0, 54.91)(1, 56.74)(2, 67.87)(3, 68.28)(4, 68.36)(5, 67.83)(6, 68.15)(7, 67.63)(8, 68.24)(9, 68.26)(10, 67.81)(11, 68.11)(12, 67.78)(13, 68.13)
    };
    \addlegendentry{RP+PCA};
  
    \addplot[
        color=cyan,
        mark=diamond,
    ]
    coordinates {
        (0, 52.34)(1, 54.87)(2, 50.26)(3, 56.89)(4, 57.83)(5, 57.29)(6, 56.52)(7, 56.16)(8, 55.73)(9, 55.64)(10, 54.63)(11, 54.27)(12, 54.88)(13, 52.60)(14, 53.55)
    };
    \addlegendentry{RP+LDA};
    \end{axis}
    \end{tikzpicture}
}
\caption{Classification accuracy of different methods on the testing set of SData}
\label{fig:classification-accuracy-sdata}
\end{figure*}

The classification accuracy of RP and FS+RP is 86.23\% and 98.97\% on the first test data respectively, where the latter are significantly higher. On the other two datasets, the SVM based on FS+RP also has higher classification accuracy than the SVM based on RP, where the classification accuracy increases by 2.79\% and 6.23\% respectively. The accuracy of SVM based on FS+RP+LDA and FS+RP+PFS is a bit lower than that of FS+RP on the first test data, with an accuracy of 98.67\% and 98.95\% respectively, while on GSE2034, FS+RP+PFS performs a little better than FS+RP with an accuracy of 61.55\%. SVM based on FS+RP obtain highest classification accuracy of 71.05\% on the third test dataset.

LDA can also help RP to find a latent space, where the classification accuracy of SVM based on RP+LDA on BC-TCGA and GSE25066 is 98.00\% and 69.21\% respectively, while the accuracy on GSE2034 is 58.87\% which is even worse than that of RP. Recent studies have indicated that classification accuracy of LDA is afflicted by two classes of samples with small inter class distance \cite{huang2002solving}. For this reason, RP+LDA deteriorates subspace generated by RP. And experiments in the next subsection also prove this point. In contrast, methods combined with FS still lead to good classification accuracy even distance between classes is very small. Due to this, RP+FS achieves better classification accuracy of 60.89\% on GSE2034 compared to RP+LDA. On all  datasets, classification accuracy of SVM based on RP is lower than that of RP+FS, because RP projects original data to subspace where samples cannot be effectively divided into two groups in some dimensions (named ``invalid dimensions'') and FS filters out features from invalid dimensions after RP.

The accuracy of PCA and RP+PCA on the testing set of BC-TCGA is 77.31\% and 76.70\% respectively, which is significantly lower than 86.23\% of RP. However, the accuracy of SVM based on RP+PCA is much lower than that of FS+RP, RP+FS and RP+LDA. Because RP and PCA are both unsupervised approaches and features generated by them may not yield better classification performance.

Classification accuracy changes with the number of selected features (or spatial dimensions). Classification accuracy of the different methods on the testing set of GSE25066 is shown in Figure \ref{fig:classification-accuracy-gse25066}. The classification accuracy of FS+RP, FS+RP+PFS, RP+PCA and RP+FS has small changes with the change of dimension, as relatively large.  There is a steady increase in accuracy of RP when the dimension goes up. According to Johnson-Lindenstrauss (JL) lemma,  pairwise distances are preserved more preciously with the dimension of subspace rises. However, for dimension below the minimal number of components to guarantee with good probability (i.e. 148 for GSE25066), JL lemma is not suitable for this case and the classification accuracy of these subspaces is determined by intrinsic structure of data. Moreover, there is a sharply increase in the accuracy of FS+RP+LDA and RP+LDA when the dimension goes up. Compared with other methods, FS+RP has highest accuracy (around 70\%) in almost all dimensions, which achieves a significant improvement of 40\% in dimension equals to 100 compared to RP. It’s interesting that the classification accuracy of FS+RP, RP+FS, and FS+RP+PFS remains 65\%-70\% even in lower dimensions. Figure \ref{fig:classification-accuracy-gse25066} also reveals that the accuracy of RP+PCA is relatively low (about 60\%) and has small changes with the increase of dimensions. This indicates that PCA cannot improve the performance of RP. The similar results also obtained on other two datasets (Figure \ref{fig:classification-accuracy-bc-tcga} and \ref{fig:classification-accuracy-gse2034}).

It’s hard to compare running time since the dimension of classifiers based on different methods is different when the highest accuracy is obtained. As shown in Figure \ref{fig:classification-accuracy-gse25066}, FS+RP, FS+RP+PFS, RP+FS, and RP+PCA have relatively stable classification accuracy when dimension is 150, while FS+RP+LDA, RP+LDA and RP have relatively low classification accuracy. And dimension equals to 1000, most methods except RP obtain stable classification accuracy. When dimension rise to 3000, all methods have stable classification accuracy. Here we compared the running time in three different dimensions: 150, 1000, and 3000 on one of datasets (GSE25066), which is displayed in Figure \ref{fig:running-time-on-gse25066}, where consuming time of each stage in dimensional reduction methods was marked with different colors.

In many scenarios, prior knowledge and empirical data can be used in feature selection for offline processing. For example, microarray data can be preprocessed. Figure \ref{fig:running-time-on-gse25066} shows that FS+RP has the least online processing time among all methods. It indicates that offline preprocessing can not only improve classification accuracy but also greatly accelerate the speed of online processing. Furthermore, RP+LDA obtains a more discriminant subspace for linear spreadable data than RP consuming almost the same time. It’s clear that the running time rises with the dimensions, and the running time of FS+RP+PFS increases the most among all  methods.

\subsection{Verification}

To further validate the performance of different methods, we repeat the above experiments on a simulation dataset: SData, which includes 100 positive samples and 100 negative samples with 10,000 features, and each feature in SData follows normal distributions: $N(0, 0.1)$ and $N(0 \pm r, 0.1)$ for positive and negative samples respectively, here $r \in [-0.125,0.125]$. The performance of all methods on SData is listed in Table \ref{tab:accuracy-comparison-on-simulated-data}. Figure \ref{fig:classification-accuracy-sdata} displays the accuracy of different methods with the change of dimension on SData. There’s a rapid increase in RP with the increase of dimension from 100 to 800. Compared with RP, FS+RP, FS+RP+PFS and RP+FS yield better classification accuracy in low-dimensional subspace. As mentioned above, LDA leads to terrible performance on two classes of samples having similar mean values. Therefore, RP+LDA and FS+RP+LDA have low accuracy on SData. Overall these methods show the similar performance on the real and the simulated datasets.

\begin{table}[ht]
    \caption{Highest classification accuracy (\%) on SData}
    \centering
    \begin{tabular*}{\linewidth}{l @{\extracolsep{\fill}} rr}
    \toprule
                       & \multicolumn{2}{c}{\textbf{SData}}\\
    \midrule
    \textbf{Methods}   & Training Set & Testing Set \\
    \midrule
    RP                 & 100.00       & 91.03 \\
    FS + RP            & 100.00       & \textbf{92.88} \\
    FS + RP + LDA      & 88.74        & 66.59 \\
    FS + RP + PFS      & 100.00       & 92.31 \\
    RP + FS            & 100.00       & 91.01 \\
    RP + PCA           & 100.00       & 68.36 \\
    RP + LDA           & 74.87        & 57.83 \\
    \bottomrule
    \end{tabular*}
\label{tab:accuracy-comparison-on-simulated-data}
\end{table}

\section{Conclusions}
\label{conclusions}

In this paper, we compared performance of different techniques based on RP, and experimental results proved that the classification accuracy of RP can be improved by combining with other dimensionality reduction methods, such FS or LDA. However, it didn't yield better classification accuracy combining RP with PCA. FS followed by RP outperforms other methods in classification accuracy on most of the datasets with relatively lower computation cost.


\section*{Conflict of interest}

The authors declare that they have no competing interests.


\section*{Funding information}

This work is partially supported by the National Key Research and Development Program of China (Grant No.2016YFC0901905), National Natural Science Foundation of China (Grant No.61471147), National High-Tech Research and Development Program (863) of China (Grant Nos.2015AA020101, 2015AA020108), Natural Science Foundation of Heilongjiang Province (Grant No. F2016016) and the Fundamental Research Funds for the Central Universities (Grant No.HIT.NSRIF.2017037).

\bibliographystyle{elsarticle-harv}
\bibliography{reference}





\end{document}